\documentclass[10pt,twocolumn,letterpaper]{article}

\usepackage{wacv}              

\usepackage{graphicx}
\usepackage{amsmath}
\usepackage{amssymb}
\usepackage{booktabs}

\usepackage{textcomp, gensymb}
\usepackage{booktabs, multirow}
\usepackage{colortbl, soul}
\usepackage{physics}
\usepackage{wrapfig, adjustbox}
\usepackage{enumitem}
\usepackage[dvipsnames,table,xcdraw]{xcolor}
\usepackage{float}
\usepackage[font=small,skip=5pt]{caption}
\usepackage[english]{babel}
\usepackage[autostyle,english=american]{csquotes}

\usepackage[accsupp]{axessibility}

\usepackage[pagebackref,breaklinks,colorlinks]{hyperref}

\usepackage[capitalize]{cleveref}
\crefname{section}{Sec.}{Secs.}
\Crefname{section}{Section}{Sections}
\Crefname{table}{Table}{Tables}
\crefname{table}{Tab.}{Tabs.}

\def\eg{\emph{e.g.}}

\def\etal{\emph{et al.}}
\def\ie{\emph{i.e.}}
\newcommand{\hlc}[2][yellow]{{%
    \colorlet{foo}{#1}%
    \sethlcolor{foo}\hl{#2}}%
}

\MakeOuterQuote{"}

\definecolor{Gray}{gray}{0.9}

\def\wacvPaperID{166} 
\def\confName{WACV}
\def\confYear{2024}

\begin{document}

\title{CAMOT: Camera Angle-aware Multi-Object Tracking}

\author{Felix Limanta\\
Tokyo Institute of Technology\\
Meguro, Tokyo, Japan\\
{\tt\small felix@ks.c.titech.ac.jp}
\and
Kuniaki Uto\\
Tokyo Institute of Technology\\
Meguro, Tokyo, Japan\\
{\tt\small uto@ks.c.titech.ac.jp}
\and
Koichi Shinoda\\
Tokyo Institute of Technology\\
Meguro, Tokyo, Japan\\
{\tt\small sinoda@c.titech.ac.jp}
}
\maketitle

\begin{abstract}
    This paper proposes CAMOT, a simple camera angle estimator for multi-object tracking to tackle two problems: 1) occlusion and 2) inaccurate distance estimation in the depth direction. Under the assumption that multiple objects are located on a flat plane in each video frame, CAMOT estimates the camera angle using object detection. In addition, it gives the depth of each object, enabling pseudo-3D MOT. We evaluated its performance by adding it to various 2D MOT methods on the MOT17 and MOT20 datasets and confirmed its effectiveness. Applying CAMOT to ByteTrack, we obtained 63.8\% HOTA, 80.6\% MOTA, and 78.5\% IDF1 in MOT17, which are state-of-the-art results. Its computational cost is significantly lower than the existing deep-learning-based depth estimators for tracking.
\end{abstract}

\section{Introduction}
\label{sec:intro}
Multi-object tracking (MOT) \cite{mot/bewley:16:sort, mot/feichtenhofer:17, mot/bergmann:19:tracktor, mot/zhang:21:fairmot, mot/zhang:22:bytetrack, mot/cao:22:oc-sort} is a task to detect and track objects in a video across space and time while maintaining consistent identities. It is utilized in several applications, such as autonomous driving and video surveillance. Its standard paradigm consists of two stages: 1) object detection \cite{det/felzenswalb:10,det/ren:15:frcnn,det/liu:16:ssd,det/redmon:16:yolo,det/zhou:19:centernet,det/ge:2021:yolox}, wherein it detects individual objects in each frame, and 2) association \cite{mot/brasol:20,mot/zhang:21:fairmot}, wherein it associates detection results over time to form a track for each object. In this paper, we focus on the application of MOT to surveillance.

MOT faces several challenges in real-world scenarios. One significant problem is that the target object is often occluded by other objects, resulting in detection failure. Another problem is that the distance between two objects cannot be precisely estimated when they are aligned in the depth direction. This may cause incorrect object associations between different frames.

These two problems can be addressed if we know the depth of each object. To this end, Khurana \etal \cite{mot/khurana:21} plugged a depth estimator based on deep learning into the MOT framework. Although it somewhat solves the occlusion problem, the imprecise distance problem still needs to be solved. In addition, the depth estimator may require significant additional computational costs.

\begin{figure}[t]
    \centering
    \includegraphics[clip, trim=0 7cm 0 0, width=0.9\linewidth]{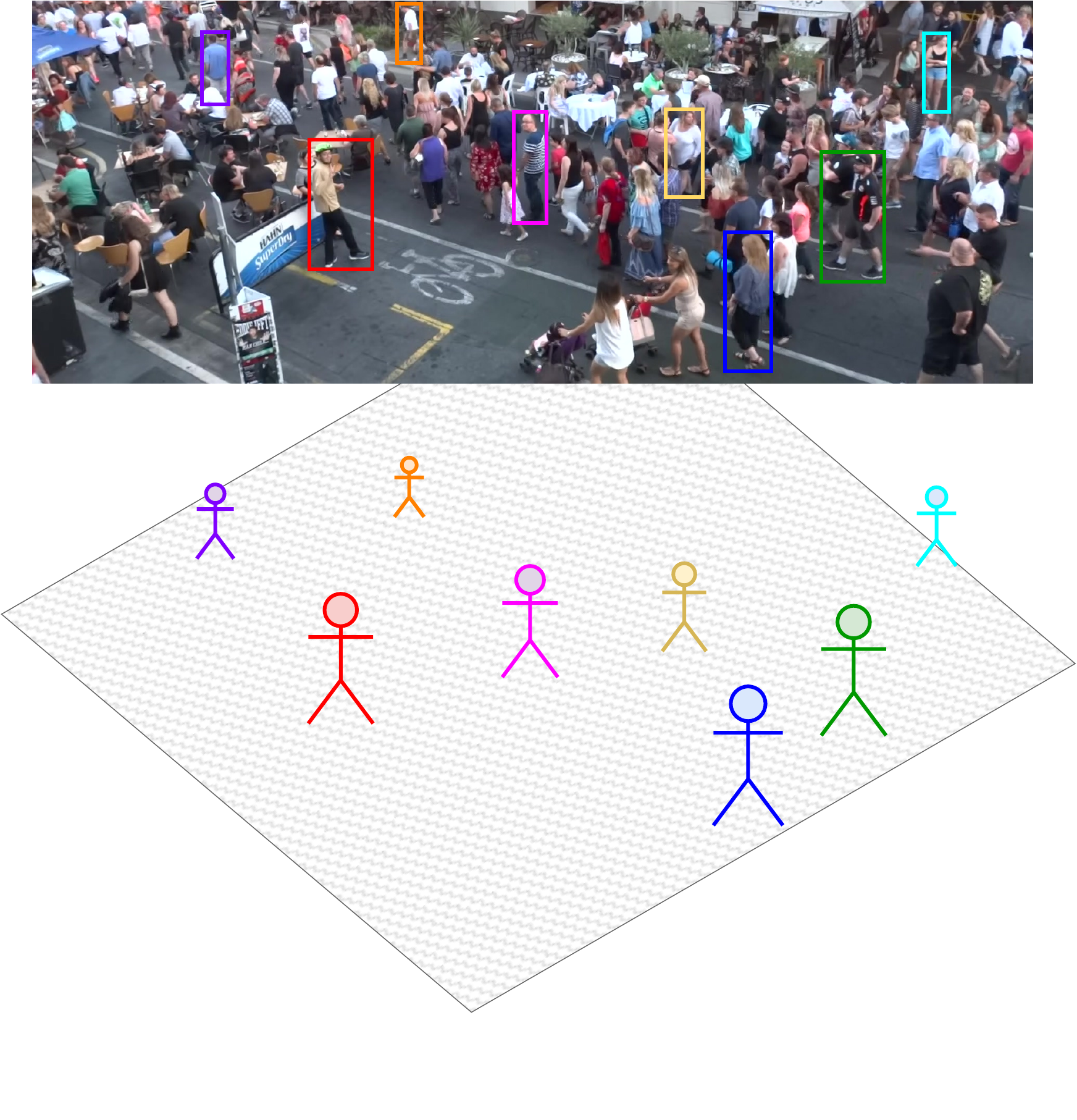}
    \caption{\textbf{Illustration on the idea of CAMOT.} Under the assumption that multiple objects are located on a flat plane, the camera angle is estimated using object detection. The scale of each bounding box indicates the depth of each object, whereas the distribution of the bounding boxes informs us of the camera angle.}
    \label{fig:intro/banner}
    \vspace{-1em}
\end{figure}

In this paper, we propose CAMOT (\textbf{C}amera \textbf{A}ngle-aware \textbf{M}ulti-\textbf{O}bject \textbf{T}racking), a simple camera angle estimator for MOT, to solve these problems. Assuming that multiple objects are located on a flat plane in each video frame, it estimates the camera angle by utilizing object detection. Our method provides the depth of each object and solves the occlusion problem. It additionally measures the distance in the depth direction and associates objects in different frames more accurately. CAMOT is computationally efficient and can be used as a plug-in component in various MOT methods.

We evaluated its performance by adding it to various 2D MOT methods on the MOT17 and MOT20 datasets and confirmed its effectiveness. For example, when applied to ByteTrack, it achieved state-of-the-art results of 63.8\% HOTA, 80.6\% MOTA, and 78.5\% IDF1 in MOT17 \cite{dataset/milan:16:mot16}. With regards to its computational cost, on a machine with a single A100 GPU, it achieved a speed of 24.92 FPS, which is higher than the sub-10 FPS speed of the existing deep-learning-based depth estimators that are used for tracking.

Overall, the main contributions of this work are summarized as follows:
\begin{enumerate}[noitemsep]
    \item We propose a lightweight camera angle estimator that leverages object detection locations.
    \item We utilize the camera angle and the depth of each object to associate objects between frames in 2D MOT.
    \item We evaluate our proposed method by adding it to various 2D MOT methods.
\end{enumerate}

\section{Related Works}
\label{sec:related_works}
\subsection{2D Multi-Object Tracking (MOT)}
With the advent of reliable object detection, the standard approach for MOT is "tracking-by-detection," which uses pre-trained detectors and focuses more on data association. Early methods such as SORT \cite{mot/bewley:16:sort} and DeepSORT \cite{mot/wojke:17:deepsort} utilize Kalman filters. In contrast, recent methods try novel approaches, such as regressing bounding boxes by frame \cite{mot/bergmann:19:tracktor}, matching heatmaps in a receptive field \cite{mot/zhou:20:centertrack}, or combining detection and reidentification (Re-ID) in a single model \cite{mot/xu:18,mot/wang:20,mot/li:21,mot/zhang:21:fairmot}. Additionally, the Vision Transformer \cite{misc/dosovitskiy:21:vit} has also made its way into MOT \cite{mot/sun:20:transtrack,mot/zeng:22:motr,mot/xu:21,mot/cai:22:memot}, combining detection and tracking in an end-to-end manner. A notable recent development in MOT is ByteTrack \cite{mot/zhang:22:bytetrack}, which modifies the simple SORT into a two-pass algorithm that processes low-confidence bounding boxes.

One problem with MOT is that the distance between two objects cannot be precisely estimated when they are aligned in the depth direction. This is because the depth direction is flattened and combined with another direction (usually the up/down direction) when a scene is projected onto a camera. Due to perspective, objects at different depths may appear to have the same distance in the image. However, none of the present methods consider perspective distortion in spatial directions when associating objects.

The most commonly used feature for association is the Intersection-over-Union (IoU) \cite{mot/bewley:16:sort,mot/bergmann:19:tracktor,mot/zhang:22:bytetrack}, followed by appearance features \cite{mot/wojke:17:deepsort,mot/zhang:21:fairmot}. Transformer-based MOT \cite{mot/meinhardt:22:trackformer,mot/zeng:22:motr,mot/cai:22:memot} attempts to unify detection and tracking and performs the association process as part of the model. All these methods use the distance measured on the 2D image.

A recent alternative to the vanilla IoU is the Distance-IoU (DIoU) \cite{misc/zheng:20:dist-iou}, which also considers the relative positioning of objects. For two different objects $i$ and $j$, the DIoU is defined as

\vspace{-0.5em}
\begin{equation}
    \text{DIoU} = \text{IoU} - \frac{d_x^2 + d_y^2}{c_x^2 + c_y^2} ,
    \label{eq:diou}
\end{equation}
\vspace{-0.5em}

\noindent where $(d_x, d_y)$ are the horizontal and vertical distances in the image plane between the center points of $i$ and $j$, and $(c_x, c_y)$ are the width and height of the minimum bounding box that covers the bounding boxes of both $i$ and $j$. Previous studies have applied DIoU to MOT \cite{mot/liang:22,mot/yuan:22,mot/stadler:2023}. We base our own association metric on the DIoU and modify it to incorporate camera angles.

\subsection{Occlusion}
Occlusion remains a major problem in MOT. A common method to handle occlusions is reidentification (Re-ID), which is used to relink detections before and after occlusions \cite{mot/bergmann:19:tracktor, mot/zhou:20:centertrack, mot/xu:18,mot/wang:20,mot/li:21,mot/zhang:21:fairmot}. However, this is a \textit{post-hoc} reasoning on the presence of occluded objects; in an online setting, intelligent agents should reason about occluded objects \textit{before} they re-appear \cite{mot/khurana:21}. Furthermore, Re-ID cannot handle longer occlusions ($>3$ s) effectively \cite{mot/dendorfer:22:quovadis} owing to the widening gap between pre- and post-occlusion features.

A possible solution is to lift the tracking process into 3D space. It is much easier to track occluded objects in 3D because trajectories that overlap in 2D are well-separated in 3D space. The most straightforward approach to lifting 2D information to 3D is to use a monocular depth estimator.

\subsection{Depth Estimation}
\label{sec:related_works/depth}
Monocular depth estimation obtains a depth map from a single image without additional sensors or modalities. This is an ill-posed problem, as a 2D scene can be projected from infinitely many 3D scenes. Classical methods~\cite{depth/nagai:02,depth/saxena:05} rely on a thorough \textit{a priori} knowledge of a scene and/or objects in the scene. In contrast, recent deep-learning-based methods can reliably infer depth from a single image without requiring geometric constraints or \textit{a priori} knowledge. The methods range from hourglass networks \cite{depth/li:18}, encoder-decoder structures \cite{depth/yuan:22:newcrfs,depth/bhat:23:zoedepth}, transformers \cite{depth/ranftl:21:dpt,depth/kim:22:glpn,depth/agarwal:22:depthformer}, to diffusion networks \cite{depth/saxena:23:diffusion}. Depth estimation is a precursor to several tasks, including 3D detection and tracking.

Datasets for 3D tasks (\eg, depth estimation, 3D detection, and 3D tracking) are usually collected for tasks such as autonomous driving \cite{dataset/geiger:2012:kitti,dataset/sun:2020:waymo,dataset/caesar:2020:nuscenes}. As a result, there are currently no available datasets for use cases such as surveillance.

\subsection{Depth Estimation for MOT}
Plenty of work has been done on monocular 3D object detection and tracking (\ie, generating 3D bounding boxes) with only a single RGB image. Early 3D object detection methods \cite{3d/ali:18,3d/yang:19,3d/shi:22,3d/zhang:22} work on point clouds taken by additional sensors (\eg, LiDAR, time-of-flight camera, stereo camera) or generated by monocular depth estimators. In contrast, recent methods \cite{3d/brazil:19,3d/liu:20,3d/wang:21} generate 3D bounding box proposals from a single 2D image. Several 2D MOT methods \cite{mot/bewley:16:sort,mot/wojke:17:deepsort,mot/bergmann:19:tracktor,mot/zhang:22:bytetrack} can also perform 3D MOT with only slight modifications. Additionally, dedicated 3D trackers \cite{3d/weng:19,3d/chiu:21} have also been proposed.

There also exist studies that use depth information for 2D MOT. However, unlike 3D MOT, these studies still use 2D bounding boxes as input and output; however, internal processing, such as association and trajectory management, is performed in 3D. The first study to add 3D reasoning for 2D MOT was Khurana \etal \cite{mot/khurana:21}, which utilized monocular depth estimation \cite{depth/li:18} to generate a depth map and augment SORT. By modifying the Kalman filter to track in 3D, the tracker can filter false positives for forecasts that are supposed to be occluded, thus allowing for better occlusion handling. Another work, Quo Vadis \cite{mot/dendorfer:22:quovadis}, transforms the entire scene into a birds-eye view using homography transformation and then adds trajectory prediction. Depth estimators are computationally expensive \cite{depth/yuan:22:newcrfs};  Quo Vadis \cite{mot/dendorfer:22:quovadis} adds other computationally expensive components besides the depth estimator, making the system too costly to run in any practical scenario. In comparison, our proposed estimator requires much less computational cost because it uses readily available bounding boxes.

\subsection{Camera Pose Estimation Methods}
Traditional camera pose estimation methods, \eg, RANSAC \cite{misc/fischler:81:ransac} and PnP \cite{pose/lu:18}, require handcrafted features to match between images. In contrast, deep-learning-based monocular camera pose estimation methods \cite{pose/kendall:15,pose/brahmbhatt:18} (which also produce camera angles) have produced good results with only a single image. However, most camera pose estimation methods require a video from a dynamic moving camera or require a well-defined structure as a reference. Unfortunately, many MOT scenes are captured with a static camera and contain large crowds, which may obscure the reference structure.

Our proposed method estimates camera angles with object detection output without requiring a reference structure or a dynamic camera. In addition, it can estimate camera-relative 3D points of every object without any additional training.

\section{Camera Angle Estimation}
\label{sec:method}
\begin{figure}[t]
    \centering
    \includegraphics[trim=0 0.25cm 0 0, width=0.95\linewidth]{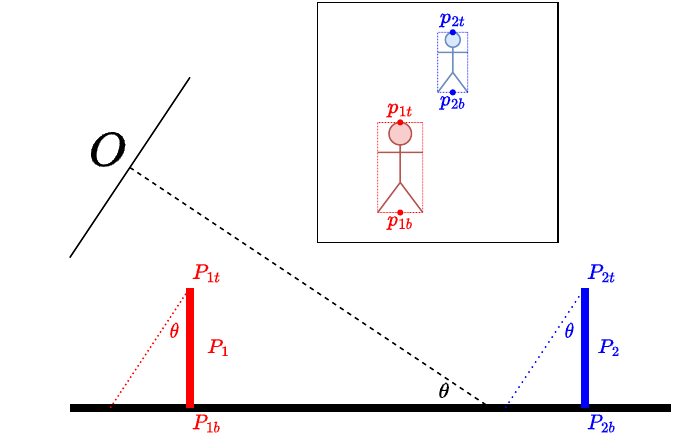}
    \caption{2D planar side view of the system. \textcolor{black}{\textbf{Black}} parts show the part of the system shared by all objects, whereas \textcolor{blue}{\textbf{blue}} and \textcolor{red}{\textbf{red}} parts show different objects.}
    \label{fig:method/system_sideview}
    \vspace{-1em}
\end{figure}

\subsection{Outline}
\label{sec:method/outline}
Here, we describe our method for estimating the camera elevation angle $\theta$ and the set of object 3D coordinates $\mathbb{P}$. CAMOT simultaneously estimates the angle and object depths by regressing a common plane for all object detections.

Object detections, among other things, inform us where objects are distributed on an image, whereas their distributions inform us of the camera angle. For example, an image taken from a ground-level angle will have its objects concentrated in a horizontal line, whereas an image with a higher angle will have its objects distributed more evenly. We can use object detections to estimate the depth of an object, which can then be used to estimate the camera angle.

An outline of our algorithm is as follows:
\begin{enumerate}[noitemsep]
    \item Select bounding boxes to use in the current frame $t$.
    \item While $\theta^t$ is not optimal $\left( \varepsilon^{(t,u)} > \tau_\varepsilon \right)$, advance the iteration $u \leftarrow u + 1$ as outlined below:
        \begin{enumerate}[noitemsep]
            \item Set a $\theta^{(t,u)}$ value for the current iteration.
            \item Estimate the 3D object points $P_i^{(t,u)}$ using $\theta^{(t,u)}$.
            \item Regress a plane with the normal vector $\vb{n}^{(t,u)}$ from $P_i^{(t,u)}$ and calculate the plane angle $\theta_{\vb{n}}^{(t,u)}$.
            \item Evaluate the angle estimation process error $\varepsilon^{(t,u)}$ for this iteration.
        \end{enumerate}
    \item Apply angle smoothing for $\theta^t$.
    \item Use the optimal $\theta^t$ value to calculate $P_i^t$ for \textit{all} objects in the current frame.
\end{enumerate}

\subsection{Assumptions and Problem Formulation}
\label{sec:method/assumption}
For our intended use case (surveillance, crowd analysis, etc.), we limit the scenario to tracking the movements of a crowd (of humans) in a public space. We set the following assumptions regarding the scenario:

\begin{enumerate}[noitemsep]
    \item The camera parameters (\eg, $\theta$) are unknown, except for the focal length $f$. \label{list:method/assumption_f}
    \item At least \textbf{three} objects are detected in each frame of the video.
    \item An object is assumed to be in contact with the ground, \ie, the bottom edges of all objects lie on a common plane (ground plane).
    \item The change of camera angles is smooth over time. \label{list:method/assumption_smooth}
\end{enumerate}

We model a human object as a cylinder in 3D space with a centroid $P_i = (X_i, Y_i, Z_i)$, height $H$, and an aspect ratio $A$. We model the height $H$ as constant, but the aspect ratio $A$ can vary. We use pinhole camera optics with a focal length $f$.

Under the pinhole camera model, we formulate the problem as finding the optimal $\theta$ value that minimizes the regression error $\varepsilon$ for the best-fit plane between all detected objects. Figure~\ref{fig:method/system_sideview} illustrates the problem, where an object $i$ is represented as a line segment $\overline{P_{it} P_{ib}}$ passing through $P_i$, with $P_{it}$, $P_i$, and $P_{ib}$ as the top, middle, and bottom points of an object, respectively.

\subsection{Bounding Box Selection}
Not all bounding boxes produced by the object detection process can be used for the plane estimation process. Let $i$ be an index for all detected objects in the current frame. We first need to select $n_{\text{plane}}$ objects, where $n_{\text{plane}}$ is the target number of objects to use in the plane estimation process.

We first filter out objects whose corresponding bounding boxes clip the edge of the frame. We then divide the image into $n_{\text{plane}}$ regions width-wise. For each region, we add to $\mathbb{I}_{\text{plane}}$ at most one object with the highest detection confidence whose centroid lies in that region. It is possible for a region to be empty (\ie, does not contain any detection). In that case, $\left| \mathbb{I}_{\text{plane}} \right|$ may be less than $n_{\text{plane}}$.

\subsection{Initial Elevation Angle Setting}
We define the camera elevation angle $\theta$ as the angle between the camera principal axis (\ie, the $z$-axis) and its projection on the ground plane. Given the set of 3D points $\mathbb{P} = \{ P_i \}$, it is trivial to obtain $\theta$ (see Section~\ref{sec:method/plane}). However, as both $\theta$ and $\mathbb{P}$ are unknown at the beginning, we first assume an initial $\theta$ value and iteratively obtain a final $\theta$ value through optimization (see Section~\ref{sec:method/error}).

In the first frame, $\theta$ is initialized according to the parameter $\theta^0$. In later frames, $\theta$ is initialized as the previous frame's value.

\subsection{Depth Estimation Using Detection Results}
\label{sec:method/depth}
\begin{figure}[t]
    \centering
    \includegraphics[trim=0 0.75cm 0 0, width=0.75\linewidth]{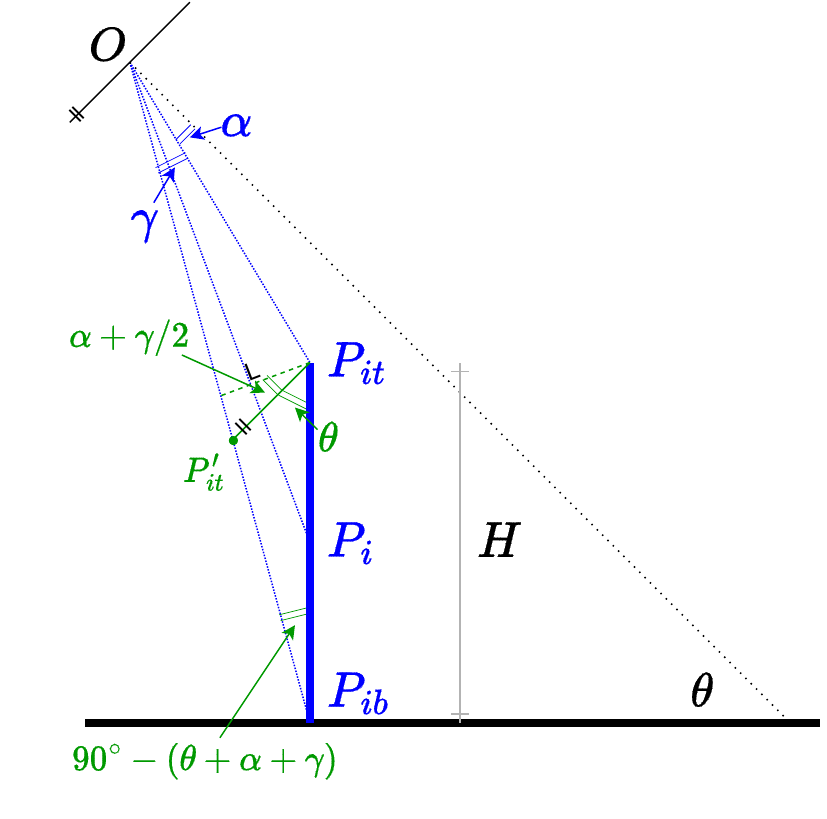}
    \caption{2D planar side view for one object. \textcolor{black}{\textbf{Black}} parts show part of the system shared by all objects, while \textcolor{blue}{\textbf{blue}} parts show components unique for the object $i$. \textcolor{OliveGreen}{\textbf{Green}} parts show derived points, angles, etc., for calculation.}
    \label{fig:method/angle_triangles}
    \vspace{-1em}
\end{figure}

This process is first performed in each optimization iteration over the selected objects $\mathbb{I}_{\text{plane}}$ to obtain the set of selected points $\mathbb{P}_{\text{plane}} = \{ P_i \}$. After the optimal value for $\theta$ is obtained, one last pass is performed over the set of all objects $\mathbb{I}$ to obtain the set of all 3D points $\mathbb{P}$.

Using the angle $\theta$, we can find the distance $d_i$ from the origin point to $P_i$, $P_{it}$, and $P_{ib}$. We use the top and bottom points for calculation since they are the nearest and furthest points an object has from the camera origin point, respectively.

We first convert 2D coordinates into 3D rays originating from the camera focal point, defined as

\vspace{-0.5em}
\begin{equation}
    r_i
    = \kappa_c^{-1}
    \begin{bmatrix}
        x_i & y_i & 1
    \end{bmatrix}^T ,
    \label{eq:angle_backproj}
\end{equation}

\noindent where the ray $r_i = (a_i, b_i, 1)$ is the ray passing through the pinhole camera origin point, the relevant point in the image plane, and the actual point in the 3D camera coordinate, and $\kappa_c^{-1}$ is the inverse of the camera intrinsic matrix. For every object $i$, we obtain the rays to the top point $r_{it}$ and the bottom point $r_{ib}$ by using the top-middle and bottom-middle 2D positions $(x_i, y_{it}), (x_i, y_{ib})$ of a bounding box, respectively, as illustrated in Fig.~\ref{fig:method/system_sideview}.

We then define the following: for each object $i$, the angle $\alpha_i$ is defined as the angle between the principal axis and the $r_{it}$, whereas the angle $\gamma_i$ is defined as the angle between $r_{it}$ and $r_{ib}$.

In Fig.~\ref{fig:method/angle_triangles}, we can use the angle properties in the triangles formed by the object and in the rays to calculate the distances from the origin point to the object. Using the triangle $\triangle O P_{it} P_{i}$, the distance to the head point $d_{it}$ can be obtained as follows.

\vspace{-0.5em}
\begin{equation}
    d_{it} = \frac{H \cos \left( \theta + \alpha_i + \frac{\gamma_i}{2} \right)}{2 \sin \frac{\gamma_i}{2}}
    \label{eq:angle_dit}
\end{equation}
\vspace{-0.5em}

\noindent Then, using $\triangle O P_{it} P_{ib}$, the distance to the bottom point $d_{ib}$ can be obtained as follows.

\vspace{-0.5em}
\begin{equation}
    d_{ib} = \frac{2 d_{it} \sin \left( \frac{\gamma_i}{2} \right) \sin \left( \theta + \alpha_i + \frac{\gamma_i}{2} \right)}{\cos \left( \theta + \alpha_i + \gamma_i \right)} + d_{it}
    \label{eq:angle_dib}
\end{equation}
\vspace{-0.5em}

Afterwards, using the rays $(r_{it}, r_{ib})$, the centroid point $P_i$ is calculated following Eq.~\ref{eq:angle_Pi}.

\vspace{-0.5em}
\begin{equation}
    P_{i} = \frac{r_{it} d_{it} \cos \alpha_i + r_{ib} d_{ib} \cos \left( \alpha_i + \gamma_i \right)}{2}
    \label{eq:angle_Pi}
\end{equation}
\vspace{-0.5em}

Applying this to the selected objects $\mathbb{I}_{\text{plane}}$, we obtain the set of 3D object points $\mathbb{P}_{\text{plane}}$ to use in the plane estimation process. In the final pass, applying this to all objects $\mathbb{I}$, we obtain the final set of all 3D object points $\mathbb{P}$.

\subsection{Plane Estimation}
\label{sec:method/plane}
$\mathbb{P}_{\text{plane}}$ is fit to a plane $(\vb{n}, P_{\text{plane}})$, with $\vb{n}$ being the plane normal unit vector and $P_{\text{plane}} = (0, 0, z_{\text{plane}})$ being the intersection of the $z$-axis and the plane.

Finally, the plane is used to obtain the angle of the plane $\theta_{\vb{n}}$. Provided that $\vb{n} = (n_x, n_y, n_z)$ is a unit vector, $\theta_{\vb{n}}$ is obtained with

\vspace{-1em}
\begin{equation}
    \theta_{\vb{n}} = \arccos(n_z)
    \label{eq:angle_theta_n}
\end{equation}

Note that plane estimation in 3D requires at least three points. If there are too few objects in the frame (\ie, $|\mathbb{P}_\text{plane}| < 3$), then we keep the plane from the previous frame (\ie, $(\vb{n}, P_{\text{plane}})^t = (\vb{n}, P_{\text{plane}})^{t-1}$)

\subsection{Error Calculation}
\label{sec:method/error}
We aim to find the elevation angle value $\theta$ that minimizes the error $\varepsilon$. We calculate three error terms: 1) the perpendicularity constraint $\varepsilon_{\vb{n}}$, in which an object should stand perpendicular to the plane, \ie, $\vb{n}$ and the vector $\vb{v}_i = P_{it} - P_{ib}$ should be parallel (Eq.~\ref{eq:angle_errplane}); 2) the plane angle error $\varepsilon_{\theta}$, which is the normalized angle difference between the assumed $\theta$ and the angle of the regressed plane $\theta_{\vb{n}}$ in radians (Eq.~\ref{eq:angle_errgrad}); and 3) the regression error $\varepsilon_{\text{regr}}$, which is the root mean squared error (RMSE) for the distance between $P_i$ and the regressed plane (Eq.~\ref{eq:angle_errregr}).

\vspace{-1em}
\begin{align}
    \varepsilon_{\vb{n}} &= \frac{1}{n} \sum_{i = 1}^{n} \left[ 1 - \frac{\vb{v}_i \cdot \vb{n}}{\left\lVert \vb{v}_i \right\rVert \left\lVert \vb{n} \right\rVert } \right]
    \label{eq:angle_errplane} \\
    \varepsilon_{\theta} &= \frac{2}{\pi} \left\lvert \theta - \theta_{\vb{n}} \right\rvert
    \label{eq:angle_errgrad} \\
    \varepsilon_{\text{regr}} &= \sqrt{\frac{1}{n} \sum_{i = 1}^{n} \left\lVert \left[ \vb{n} \cdot (P_i - P_{\text{plane}}) \right] \times \vb{n} \right\rVert^2}
    \label{eq:angle_errregr}
\end{align}

We calculate the final error $\varepsilon$ as a weighted sum of the three error terms:

\vspace{-1em}
\begin{equation}
    \varepsilon = \lambda_{\vb{n}} \varepsilon_{\vb{n}} + \lambda_{\theta} \varepsilon_{\theta} + \lambda_{\text{regr}} \varepsilon_{\text{regr}}
    \label{eq:angle_err}
\end{equation}

We use the Nelder--Mead optimization algorithm \cite{misc/nelder:65} with the error $\varepsilon$ as the minimization objective when finding the optimal angle. The optimization process stops when $\varepsilon$ is smaller than the convergence threshold $\tau_\varepsilon$.

The final product is the optimal elevation angle $\theta$.

\subsection{Angle Smoothing}
To enforce the assumption that depth estimates (and thus angles) are smooth over time, we apply weighted moving average smoothing with a window of $w = \text{fps} / 2$.

\begin{equation}
    \theta^t = \frac{2}{w(w + 1)} \sum_{i = t - w + 1}^{t} \left[ (i - t + w) \theta^i \right]
    \label{eq:moving_avg}
\end{equation}

Using the smoothed angle $\theta^t$ for frame $t$, we recalculate the set of all points $\mathbb{P}^t$ to obtain the final representation for frame $t$.

The camera angles and object depths produced may not be accurate with the ground truth; however, as long as the angles and depths are consistent (\ie, smooth over time) throughout the video, our method does not require them to be accurate.

\section{Angle- and Depth-aware MOT}
\subsection{Tracking in 3D Camera Coordinates}
Many MOT methods \cite{mot/bewley:16:sort,mot/wojke:17:deepsort,mot/zhang:22:bytetrack} employ the Kalman filter to predict the location of objects and reconcile it with the detected object positions. The state space of a Kalman filter used in tracking typically includes the position $(x, y)$, aspect ratio $(a)$, height $(h)$, and their velocities $(\dot{x}, \dot{y}, \dot{a}, \dot{h})$. We add an inverse depth term $1/z$ to this state space, which gives the state vector $\left( x_i, y_i, \frac{1}{z_i}, a_i, \dot{x}_i, \dot{y}_i, \dot{h}_i \right)$ \cite{mot/khurana:21}. The values of $(x_i, y_i, z_i)$ are obtained from the 3D centroid point $P_i$ produced in Section~\ref{sec:method/depth}.

To account for variations in depth, following \cite{mot/khurana:21}, we scale the Gaussian noise by the inverse depth, resulting in a constant velocity model, as shown in Eq. \ref{eq:trk_velocity_model}. This leads to smoother tracks for objects that are far away.

\vspace{-0.5em}
\begin{equation}
    x_t \approx x_{t-1} + \dot{x}_{t-1} + f\frac{\epsilon_x}{z_{t-1}}
    \label{eq:trk_velocity_model}
\end{equation}

To account for camera motion, following \cite{mot/bergmann:19:tracktor,mot/khurana:21}, we apply camera motion compensation (CMC) by aligning frames via image registration using the Enhanced Correlation Coefficient (ECC) maximization introduced in \cite{misc/evangelidis:08}.

\subsection{Track Association}

To account for the camera elevation angle, we modify the similarity matrix $K = [k_{ij}]$ based on the Distance-IoU \cite{misc/zheng:20:dist-iou}. From the original formula in Eq~\ref{eq:diou}, we add the camera angle factor $\phi$ to the y-axis terms ($d_y$ and $c_y$). The modified similarity measure is defined as

\vspace{-1em}
\begin{align}
    k'_{ij} &= k_{ij} - \frac{d_x^2 + \phi d_y^2}{c_x^2 + \phi c_y^2} ,
    \label{eq:trk_assoc_angle} \\
    \phi &= 1 + \cos^2 \theta , \label{eq:trk_assoc_angle_factor}
\end{align}
\vspace{-1em}

\noindent where $i$ and $j$ are the detection and track, respectively, $k_{ij}$ is the original similarity measure (2D IoU), and $k'_{ij}$ is the modified similarity measure (DIoU with angles).

Low elevation angles cause $\phi$ to have a large value and $k'_{ij}$ a smaller value, thereby reducing the similarity between two objects if they are positioned above or below each other. This discourages searching in the image's vertical direction, as objects typically do not move vertically when the camera is at ground level. For high elevation angles, where objects have more freedom of movement around the vertical axis, $\phi$ has less effect and Eq.~\ref{eq:trk_assoc_angle} degrades to the normal DIoU similarity.

The resulting similarity matrix $K' = [k'_{ij}]$ incorporates camera angle information and is used in the object association step.

\subsection{Assumptions and Limitations}
CAMOT also relies on the assumptions stated in Section~\ref{sec:method/assumption}. Particularly, for tracking in 3D coordinates, we assume that the camera focal length $f$ is known (Assumption~\ref{list:method/assumption_f}). In practical applications, it is possible to calibrate the camera to satisfy this assumption. However, most videos found on the Internet (\eg, YouTube, etc.), including the videos in MOT evaluation datasets, do not come with camera intrinsics. Thus, we tune the value of $f$ on the training set and select a single $f$ value that best fits our dataset. We observe that the value of $f$ can be generalized sufficiently for typical video sequences.

Another important assumption is that the change of camera angles (and thus depth estimates) is smooth over time (Assumption \ref{list:method/assumption_smooth}).
Once again, our method works even though the values are not accurate, as long as they are consistent.

In addition, CAMOT benefits when more objects are detected in the current frame, as the plane estimation would be more stable. However, as mentioned in Section~\ref{sec:method/plane}, CAMOT only works if there is a minimum of three objects at any time in the frame.

Although these assumptions and limitations may not always hold in real-world scenarios, our empirical results show that our method is applicable to different scenarios.

\section{Experiments}
\label{sec:experiments}
\subsection{Experiment Setup}
\noindent \textbf{Datasets.} We evaluate our method on the MOTChallenge \cite{dataset/milan:16:mot16,dataset/dendorfer:20:mot20} (\ie, MOT17 and MOT20) datasets. As standard protocols, CLEAR MOT Metrics \cite{dataset/milan:16:mot16} and HOTA \cite{dataset/luiten:21:hota} are used for evaluation.

\vspace{1em}
\noindent \textbf{Implementation details.} We implemented our proposed method in PyTorch \cite{misc/paszke:19:pytorch} and performed all experiments on a system with 8 NVIDIA Tesla A100 GPUs. We used ByteTrack \cite{mot/zhang:22:bytetrack} as a baseline and built our method on its top. Most hyperparameters that are used for tracking are kept the same, with the detection thresholds $\tau_{\text{high}}, \tau_{\text{low}}$ kept as $0.6$ and $0.2$, respectively. Lost tracklets were kept for 30 frames before being discarded.

For object detection, we used YOLOX-x \cite{det/ge:2021:yolox} pre-trained in COCO. The model was trained on a mix of MOT17, MOT20, CrowdHuman, Cityperson, and ETHZ for 80 epochs with a batch size of 48. We used SGD as an optimizer with a weight decay of $5 \times 10^{-4}$ and a momentum of $0.9$. The initial learning rate is $10^{-3}$ with a 1 epoch warm-up and cosine annealing schedule.

For angle estimation, we excluded all objects with bounding boxes partially outside the frame. We set $f = 50$mm and $n_{\text{plane}} = 40$. For objects clipped on the top or bottom edges, we used the 3D object height $H$ defined in \ref{sec:method/assumption} to extrapolate the clipped side in relation to the visible side. For objects clipped on the left or right edges, we extrapolated the clipped bounding box points using the average aspect ratio. We empirically set the initial input angle $\theta^0$ to $0\degree$, the error weights $[\lambda_{\vb{n}}, \lambda_{\theta}, \lambda_{\text{regr}}]$ to $[0.6, 0.3, 0.1]$, and the convergence threshold $\tau_\varepsilon$ to $10^{-4}$.

Following \cite{det/ge:2021:yolox}, FPS was measured using FP16 precision \cite{misc/micikevicius:18} and a batch size of 1 on a single GPU. We use a machine running an AMD EPYC 7702 1.5GHz with 256GiB RAM and one NVIDIA A100 GPU.

\subsection{Results}
\noindent \textbf{Benchmarks.} Table~\ref{tab:exp_mot17} and Table~\ref{tab:exp_mot20} present a comparison of our tracker with the other mainstream MOT methods on the test sets of MOT17~\cite{dataset/milan:16:mot16} and MOT20~\cite{dataset/dendorfer:20:mot20}, respectively. Since detection quality significantly affects overall tracking performance, for a fair comparison, the methods in the lower block use the same detections generated by YOLOX~\cite{det/ge:2021:yolox}, with YOLOX weights for the MOT17 and MOT20 datasets provided by ByteTrack~\cite{mot/zhang:22:bytetrack} and OC-SORT~\cite{mot/cao:22:oc-sort}, respectively. We also list the methods in the top block for reference, which may use better or worse detections than ours.

Our method outperforms all previous approaches in HOTA, MOTA, and IDF1, while being slightly inferior on FP, FN, and IDSw. On MOT17, our method exhibits the least identity switch (IDSw) errors compared with other methods using the same detection, whereas on MOT20, we narrowly come second. Our method improves on the original ByteTrack~\cite{mot/zhang:22:bytetrack} baseline and generally outperforms all other methods.

Compared to the baseline (ByteTrack~\cite{mot/zhang:22:bytetrack}), we have obtained less \#FP and \#IDSw, but introduced more \#FN. The 3D tracking approach we employ results in better separation of trajectories, leading to a smaller amount of viable detection-track pair candidates.

\begin{table*}[t]
    \centering
        \begin{tabular}{@{}l|rrrrrr@{}}
            \toprule
            \textbf{Method} &
            \multicolumn{1}{c}{\textbf{HOTA $\uparrow$}} &
            \multicolumn{1}{c}{\textbf{MOTA $\uparrow$}} &
            \multicolumn{1}{c}{\textbf{IDF1 $\uparrow$}} &
            \multicolumn{1}{c}{\textbf{FP $(10^4)$ $\downarrow$}} &
            \multicolumn{1}{c}{\textbf{FN $(10^4)$ $\downarrow$}} &
            \multicolumn{1}{c}{\textbf{IDSw $\downarrow$}} \\

            \midrule
            FairMOT~\cite{mot/zhang:21:fairmot} & 59.3 & 73.7 & 72.3 & 2.75 & 11.7 & 3303 \\
            GRTU~\cite{mot/wang:21:grtu} & 62.0 & 74.9 & 75.0 & 3.20 & 10.8 & \textbf{1812} \\
            MOTR~\cite{mot/zeng:22:motr} & 57.2 & 71.9 & 68.4 & 2.11 & 13.6 & 2115 \\
            TransMOT~\cite{mot/chu:23:transmot} & 61.7 & 76.7 & 75.1 & 3.62 & 9.32 & 2346 \\
            MeMOT~\cite{mot/cai:22:memot} & 56.9 & 72.5 & 69.0 & 2.72 & 11.5 & 2724 \\
            UniCorn~\cite{mot/yan:22:unicorn} & 61.7 & 77.2 & 75.5 & 5.01 & \textbf{7.33} & 5379 \\
            \rowcolor{Gray}
            ByteTrack~\cite{mot/zhang:22:bytetrack} & 63.1 & 80.3 & 77.3 & 2.55 & 8.37 & 2196 \\
            \rowcolor{Gray}
            OC-SORT~\cite{mot/cao:22:oc-sort} & 63.2 & 78.0 & 77.5 & \textbf{1.51} & 10.8 & 1950 \\
            \rowcolor{Gray}
            CAMOT (ours) & \textbf{63.8} & \textbf{80.6} & \textbf{78.5} & 1.85 & 8.96 & 1843 \\ 
            \bottomrule
        \end{tabular}
    \caption{Results on the MOT17-test dataset with private detections. Methods in the \hlc[Gray]{gray} block share the same detections.}
    \label{tab:exp_mot17}
    \vspace{-1em}
\end{table*}

\begin{table*}[t]
    \centering
        \begin{tabular}{@{}l|rrrrrr@{}}
            \toprule
            \textbf{Method} &
            \multicolumn{1}{c}{\textbf{HOTA $\uparrow$}} &
            \multicolumn{1}{c}{\textbf{MOTA $\uparrow$}} &
            \multicolumn{1}{c}{\textbf{IDF1 $\uparrow$}} &
            \multicolumn{1}{c}{\textbf{FP $(10^4)$ $\downarrow$}} &
            \multicolumn{1}{c}{\textbf{FN $(10^4)$ $\downarrow$}} &
            \multicolumn{1}{c}{\textbf{IDSw $\downarrow$}} \\

            \midrule
            FairMOT~\cite{mot/zhang:21:fairmot} & 54.6 & 61.8 & 67.3 & 10.3 & 8.89 & 5243 \\
            TransMOT~\cite{mot/chu:23:transmot} & 61.9 & 77.5 & 75.2 & 3.42 & \textbf{8.08} & 1615 \\
            MeMOT~\cite{mot/cai:22:memot} & 54.1 & 63.7 & 66.1 & 4.79 & 13.8 & 1938 \\
            \rowcolor{Gray}
            ByteTrack~\cite{mot/zhang:22:bytetrack} & 61.3 & 77.8 & 75.2 & 2.62 & 8.76 & 1223 \\
            \rowcolor{Gray}
            OC-SORT~\cite{mot/cao:22:oc-sort} & 62.1 & 75.5 & 75.9 & \textbf{1.80} & 10.8 & \textbf{913} \\
            \rowcolor{Gray}
            CAMOT (ours) & \textbf{62.8} & \textbf{78.2} & \textbf{76.1} & 2.09 & 9.13 & 945 \\ 
            \bottomrule
        \end{tabular}
    \caption{Results on the MOT20-test dataset with private detections. Methods in the \hlc[Gray]{gray} block share the same detections.}
    \label{tab:exp_mot20}
\end{table*}

\vspace{1em}
\noindent \textbf{Inference speed.} We also demonstrate that our method can operate in real time without significantly reducing inference speed. As shown in Table~\ref{tab:exp_fps}, although incorporating angle and 3D point estimation do have a slight impact on speed, it is not significant. The benefits of improved performance more than offset any modest decrease in speed.


\begin{table}[t]
    \centering
    \begin{tabular}{@{}l|r@{}}
        \toprule
        \textbf{Method} &
        \multicolumn{1}{c}{\textbf{FPS $\uparrow$}} \\

        \midrule
        TrackFormer~\cite{mot/meinhardt:22:trackformer} & 10.0 \\
        ByteTrack~\cite{mot/zhang:22:bytetrack} & \textbf{29.6} \\
        CAMOT (ours) & 27.9 \\
        \bottomrule
    \end{tabular}
    \caption{Comparison of the detection/tracking speeds generated by CAMOT and other existing methods. For CAMOT, all components listed in the ablation study (Table~\ref{tab:exp_ablation}) are active here.}
    \label{tab:exp_fps}
\end{table}

\subsection{Ablation Studies}
\noindent \textbf{Component analysis.} Table~\ref{tab:exp_ablation} shows the results of the ablation studies conducted using the MOT17 dataset. We test for four variables: whether angle estimation is performed on the entire video or just on the first frame (Var$\theta$); whether the depth forecast is used in the Kalman Filter (DF); whether the angle-aware association is performed (AA); and whether camera motion compensation (CMC) is used. The results show that each component improves tracking performance. In addition, we note that CMC significantly improves the performance, perhaps due to the number of moving camera sequences in the MOT17 dataset.

\begin{table}[t]
    \centering
    \begin{tabular}{@{}cccc|rr@{}}
        \toprule
        \multicolumn{1}{c}{Var$\theta$} &
        \multicolumn{1}{c}{DF} &
        \multicolumn{1}{c}{AA} &
        \multicolumn{1}{c|}{CMC} &
        \multicolumn{1}{c}{\textbf{MOTA $\uparrow$}} &
        \multicolumn{1}{c}{\textbf{IDF1 $\uparrow$}} \\

        \midrule
        \checkmark &            &            &            & 73.2 & 74.3 \\
                   & \checkmark &            &            & 72.9 & 72.3 \\
        \checkmark & \checkmark &            &            & 74.4 & 76.1 \\
        \checkmark &            & \checkmark &            & 73.9 & 74.2 \\
        \checkmark & \checkmark & \checkmark &            & 76.2 & 79.2 \\
        \checkmark & \checkmark & \checkmark & \checkmark & 78.4 & 81.2 \\
        \bottomrule
    \end{tabular}
    \caption{Ablation study on the MOT17 validation set. Var$\theta$ indicates whether angle estimation is performed on the entire video or just on the first frame; DF indicates whether the depth forecast is used in the Kalman Filter; AA indicates whether the angle-aware association is performed; and CMC indicates whether camera motion compensation is used.
    }
    \label{tab:exp_ablation}
\end{table}

\vspace{1em}
\noindent \textbf{Per-sequence analysis.} To test the effect of camera angles on tracking performance, based on the qualitative camera angle in the first frame, we divided the MOT17 validation set into low-angle $(\theta \leq 15\degree)$ and high-angle $(\theta > 15\degree)$ sequences. The results are shown in Table~\ref{tab:exp_angle_div}.

In lower-angle scenarios, both tracking in 3D camera coordinates and the angle-aware association significantly impact the tracking performance. By operating in 3D camera coordinates, the tracker better understands object trajectories, mitigating identity switches and improving accuracy, particularly in occlusion cases. The angle-aware association further enhances the tracking process by discouraging unlikely trajectory associations based on the current camera angle.

The angle-aware association module becomes less influential in higher-angle scenarios than in lower-angle cases. However, tracking in 3D camera coordinates remains highly effective in higher-angle scenarios. The elevated camera angle reduces occlusion and perspective distortion caused by objects in the foreground, providing a wider field of view and improved visibility.

\begin{table}[t]
    \centering
        \begin{tabular}{@{}l|rr@{}}
            \toprule
            \textbf{Sequence} &
            \multicolumn{1}{c}{\textbf{MOTA $\uparrow$}} &
            \multicolumn{1}{c}{\textbf{IDF1 $\uparrow$}} \\

            \midrule
            MOT17-02 & 46.9 $(+4.1)$ & 58.2 $(+3.2)$ \\
            MOT17-05 & 78.1 $(+1.4)$ & 78.3 $(+2.1)$ \\
            MOT17-09 & 82.8 $(+2.2)$ & 79.8 $(+2.0)$ \\
            MOT17-10 & 68.3 $(-1.7)$ & 69.7 $(+0.9)$ \\
            MOT17-11 & 70.4 $(+0.9)$ & 72.8 $(+1.0)$ \\
            \rowcolor{Gray}
            MOT17-04 & 89.7 $(+4.8)$ & 92.2 $(+2.1)$ \\
            \rowcolor{Gray}
            MOT17-13 & 78.9 $(+1.7)$ & 83.2 $(+1.4)$ \\

            \midrule
            \textbf{Overall} & 78.4 $(+2.2)$ & 81.2 $(+2.4)$  \\
            \bottomrule
        \end{tabular}
    \caption{Per-sequence analysis on the MOT17 validation set. Sequences in the white block are low-angle $(\theta \leq 15\degree)$, whereas those in the \hlc[Gray]{gray} block are high-angle $(\theta > 15\degree)$.}
    \label{tab:exp_angle_div}
\end{table}

\vspace{1em}
\noindent \textbf{Using depth estimators for tracking.}
We also tried replacing our proposed bounding-box-based depth estimation method with several off-the-shelf monocular depth estimators to evaluate its performance as a tracking component.

\begin{table}[t]
    \centering
        \begin{tabular}{@{}l|rrr@{}}
            \toprule
            \textbf{Method} &
            \multicolumn{1}{c}{\textbf{MOTA $\uparrow$}} &
            \multicolumn{1}{c}{\textbf{IDF1 $\uparrow$}} &
            \multicolumn{1}{c}{\textbf{FPS $\uparrow$}} \\

            \midrule
            DPT~\cite{depth/ranftl:21:dpt} & 74.4 & 76.2 & 4.75 \\
            GLPN~\cite{depth/kim:22:glpn} & 76.3 & 77.2 & 6.24 \\
            DepthFormer~\cite{depth/agarwal:22:depthformer} & 78.2 & 80.3 & 8.44 \\
            NewCRFs~\cite{depth/yuan:22:newcrfs} & 79.1 & 81.6 & 7.12 \\
            ZoeDepth~\cite{depth/bhat:23:zoedepth} & \textbf{80.2} & \textbf{82.3} & 8.24 \\
            \midrule
            Baseline & 78.4 & 81.2 & \textbf{24.92} \\
            \bottomrule
        \end{tabular}
    \caption{Results of replacing depth estimation with existing monocular depth estimators on the MOT17 validation set. "Baseline" refers to our bounding-box-based method.}
    \label{tab:exp_depth_add}
\end{table}

Table~\ref{tab:exp_depth_add} presents the evaluation results of the modified trackers on the MOT17 validation set. Our proposed depth estimation algorithm outperforms early depth estimators but falls short of more state-of-the-art methods. Early depth estimators often group crowds as a homogeneous blob, whereas later depth estimators exhibit some ability to handle such scenarios. Despite relying solely on bounding box location information, our method remains competitive as a tracking component and offers significantly faster processing speeds.

We hypothesize that the current limitations of monocular depth estimators in handling crowded scenes arise from the training data's incompatibility. These depth estimators are typically trained on datasets designed for autonomous driving, primarily consisting of ground-level camera views capturing vehicles and pedestrians. Consequently, they are less suited to handle high-angle scenarios characterized by densely crowded pedestrians effectively. Utilizing a depth estimator trained on a more suitable dataset with high-angle crowds and proper 3D labeling would likely improve performance.

\vspace{1em}
\noindent \textbf{Application on various MOT methods.}
To evaluate the versatility of our angle estimation method, we applied it to several state-of-the-art MOT methods, including
JDE \cite{mot/wang:20}, FairMOT \cite{mot/zhang:21:fairmot}, CenterTrack \cite{mot/zhou:20:centertrack},
and OC-SORT \cite{mot/cao:22:oc-sort}
We used the object detections produced by each tracker and applied our angle estimation method to estimate camera-relative 3D coordinates in the Kalman Filter update step. We also performed angle-aware association, similar to our method, while keeping other settings, such as training datasets and hyperparameters, the same.

\begin{table}[t]
    \centering
        \begin{tabular}{@{}l|rr@{}}
            \toprule
            \textbf{Method} &
            \multicolumn{1}{c}{\textbf{MOTA $\uparrow$}} &
            \multicolumn{1}{c}{\textbf{IDF1 $\uparrow$}} \\

            \midrule
            JDE~\cite{mot/wang:20} & 60.6 $(+0.6)$ & 65.1 $(+1.5)$ \\
            FairMOT~\cite{mot/zhang:21:fairmot} & 70.3 $(+1.2)$ & 73.2 $(+0.4)$ \\
            CenterTrack~\cite{mot/zhou:20:centertrack} & 67.4 $(+1.3)$ & 67.3 $(+3.1)$ \\
            OC-SORT~\cite{mot/cao:22:oc-sort} & 78.8 $(+0.8)$ & 78.1 $(+0.6)$ \\
            \bottomrule
        \end{tabular}
    \caption{Results of adding angle estimation and 3D association to other state-of-the-art trackers on the MOT17 validation set.}
    \label{tab:exp_3d_add}
\end{table}

Table~\ref{tab:exp_3d_add} presents the evaluation results of the modified trackers on the MOT17 validation set. The results show that adding angle estimation and 3D modeling improved the tracking performance of each method. These results demonstrate the potential of our method to be integrated with existing trackers.

\section{Conclusion}
\label{sec:conclusion}
This paper introduces CAMOT, an angle estimator for MOT. By estimating the camera angle, the tracker employs a heuristic to adapt the tracking behavior against the perspective distortion of how objects move relative to the camera. In addition, the calculated object depths also enable pseudo-3D MOT. Applying CAMOT to other 2D MOT trackers, the evaluation results on the MOT17 and MOT20 datasets demonstrate how CAMOT provides performance gains over existing methods and achieves state-of-the-art results. CAMOT is also more computationally efficient than deep-learning-based monocular depth estimators that are used for tracking.

Currently, CAMOT uses only a single frame as input when estimating the camera angle. Our future work will focus on using multiple frames to estimate the camera angle for improved stability.
We are also interested in applying CAMOT to general depth estimation problems, where we can safely assume that the room geometry and the sizes of objects are fixed.

{\small
\bibliographystyle{ieee_fullname}
\bibliography{egbib}
}

\end{document}